\documentclass[letterpaper, 10 pt, conference]{ieeeconf}

\usepackage{graphicx}
\usepackage{booktabs}
\usepackage{adjustbox}
\usepackage{wrapfig}
\usepackage[labelfont=bf,figurename=Fig.,font=small]{caption}
\usepackage{subcaption}
\usepackage{lipsum}
\usepackage{amssymb}
\usepackage{amsmath}
\usepackage{siunitx}
\usepackage{mathtools}
\usepackage{xcolor}
\usepackage[colorlinks = true,
            linkcolor = blue,
            urlcolor  = blue,
            citecolor = blue,
            anchorcolor = blue]{hyperref}
\let\labelindent\relax
\usepackage[inline]{enumitem}

\usepackage[ruled,vlined,linesnumbered]{algorithm2e}
\usepackage[noend]{algpseudocode}

\usepackage[ruled,vlined,linesnumbered]{algorithm2e}
\usepackage{algpseudocode}

\usepackage{algcompatible}
\usepackage{sidecap}
\sidecaptionvpos{figure}{t}

\usepackage{ifthen}
\usepackage{dsfont}
\usepackage[backend=biber,style=numeric-comp,sorting=none,maxbibnames=99]{biblatex}
\usepackage{balance}
\usepackage{dirtytalk}
\usepackage{afterpage}

\usepackage{times}

\allowdisplaybreaks

\newcommand{\accprob}{a}

\newcommand{\E}{\mathbb{E}}

\newcommand{\flowatacc}{b}

\newcommand{\N}{\mathbb{N}}

\newcommand{\pmin}{p_1}
\newcommand{\pmax}{p_2}
\newcommand{\Prob}{\mathbb{P}}

\newcommand{\ra}{\rightarrow}
\newcommand{\R}{\mathbb{R}}

\newcommand{\Tcutoff}{T_c}
\newcommand{\Thorizon}{T_h}

\newtheorem{remark}{Remark}

\newtheorem{proposition}{Proposition}
\newtheorem{definition}{Definition}

\newtheorem{theorem}{Theorem}

\usepackage{todonotes}

\bibliography{references}

\IEEEoverridecommandlockouts
\title{\LARGE \bf Towards Dynamic Causal Discovery with Rare Events: A Nonparametric Conditional Independence Test
}

\author{Chih-Yuan Chiu$^{1}$, Kshitij Kulkarni$^{1}$, and Shankar Sastry$^{1}$
\thanks{$^{1}$Department of Electrical Engineering and Computer Sciences, University of California, Berkeley, CA 94720 (emails: \texttt{\{chihyuan\_chiu, kshitijkulkarni, sastry\} at berkeley dot edu}).}%
}

\begin{document}

\maketitle

\thispagestyle{empty}
\pagestyle{empty}




\begin{abstract}
Causal phenomena associated with rare events occur across a wide range of engineering problems, such as risk-sensitive safety analysis, accident analysis and prevention, and extreme value theory. However, current methods for causal discovery are often unable to uncover causal links, between random variables in a dynamic setting, that manifest only when the variables first experience low-probability realizations. To address this issue, we introduce a novel statistical independence test on data collected from time-invariant dynamical systems in which rare but consequential events occur. In particular, we exploit the time-invariance of the underlying data to construct a superimposed dataset of the system state before rare events happen at different timesteps. We then design a conditional independence test on the reorganized data. We provide sample complexity bounds for the consistency of our method, and validate its performance across various simulated and real-world datasets, including incident data collected from the Caltrans Performance Measurement System (PeMS).
\end{abstract}













\section{INTRODUCTION}
\label{sec: Introduction}

The occurrence of rare yet consequential events during the evolution of a dynamical system is ubiquitous in many fields of engineering and science.  Examples include natural disasters, vehicular accidents, and stock market crashes. When studying such phenomena, it is crucial to understand the causal links between the disruptive event and the underlying system dynamics. In particular, if certain values of the system state increase the probability that the disruptive event occurs, control strategies should be implemented to steer the state away from such values. This can be accomplished, for instance, by incorporating a description of this causal relationship into the cost function that generates these control inputs in an optimization-based control framework. In general, it is important to consider the following question:

\textbf{Main Question (Q):} Given a rare event associated with the evolution of a dynamical system, does the onset of the event become more likely when the system state assumes certain values?

Below, we present a running example, invoked throughout ensuing sections to provide context.

\textbf{Running Example}: Consider the task of reducing the number of vehicular accidents on a road by identifying their causes. 
In particular, consider the scenario in which the amount of traffic on a network of roads has a causal effect on accident occurrence. For example, on busy streets, high traffic flow may render chain collisions more likely. In this case, since steady-state flows in a traffic network can be controlled via tolling, regulators can adjust the toll on each network link to redistribute flow and reduce the number of accidents that transpire \cite{Maheshwari2022DynamicTollingforInducingSociallyOptimalTrafficLoads, Maheshwari2022InducingSocialOptimalityinGamesviaAdaptiveIncentiveDesign}. Conversely, on other roads, low traffic flow may incentivize drivers to exceed the speed limit and create more opportunities for accidents to occur. In this case, traffic engineers can enforce speed limits more stringently at times of low traffic flow.

Although many well-established methods in the causal discovery literature can efficiently learn causal relationships from data, most only apply to data generated from probability distributions associated with static, acyclic Bayesian networks \cite{Glymour2019ReviewofCausalDiscoveryMethodsBasedonGraphicalModels, pearl2009causality}. Moreover, most causal discovery algorithms developed for time series data rely on stringent assumptions, such as linear dynamics and additive Gaussian noise models, or aggregate data along slices of fixed time indices \cite{Glymour2019ReviewofCausalDiscoveryMethodsBasedonGraphicalModels, gnecco2021causal, Granger1969InvestigatingCausalRelationsbyEconometricModelsandCrossSpectralMethods, 
PerezAriza2012CausalDiscoveryofDBNs}. However, rare events often occur sparsely at any fixed time and cannot be easily modeled using linear dynamics. 

To address these shortcomings, we present a novel approach for aggregating and analyzing time series data recording sparsely occurring, but consequential events, in which data is collected in a time-ordered fashion from a dynamical system. 
Our method rests on the observation that, whereas a rare event may be highly unlikely to occur \textit{at any fixed time} $t$, the probability of the event occurring \textit{at some time along the entire horizon} of interest is often much higher. Thus, we aggregate the time series data along the times of the event's first occurrence. This renders the
dataset more informative, by better representing the rare events of interest. Next, we present an algorithm that uses the curated data to analyze the causal relationships governing the occurrence of the rare event. We formally pose the question of whether the system state causally affects the occurrence of the rare event as a binary hypothesis test, with the null hypothesis $H_0$ corresponding to the negative answer, and the alternative hypothesis $H_1$ corresponding to the positive one. We prove that our proposed method is \textit{consistent against all alternatives} \cite{Lehmann1951ConsistencyAndUnbiasednessOfCertainNonparametricTests}. In other words, if $H_0$ were true, then as the number of data trajectories $N$ in the dataset approaches infinity, our approach would reject $H_1$ with probability 1. We validate the performance of our algorithm on simulated and on publicly available traffic and incident data collected from the Caltrans Performance Measurement System (PeMS). 



\section{RELATED WORK}
\label{sec: Related Work}


\subsection{Causal Discovery for Static and Time Series Data} 

Causal discovery algorithms identify causal links among a collection of random variables from a dataset of their realizations.
Common approaches include constraint-based methods (which use statistical independence tests), score-based methods (which pose causal discovery as an optimization problem), and hybrid methods
\cite{Glymour2019ReviewofCausalDiscoveryMethodsBasedonGraphicalModels, pearl2009causality, PetersJanzingSchoelkopf2017ElementsofCausalInference}.
However, most of these approaches apply only to non-temporal settings.
For time series data, Granger causality uses vector autoregression to study whether one time series can be used to predict another \cite{Granger1969InvestigatingCausalRelationsbyEconometricModelsandCrossSpectralMethods}. Other methods aggregate different data trajectories by matching time indices \cite{PerezAriza2012CausalDiscoveryofDBNs, entner2010causal}, or directly solve a time-varying causal graph \cite{Malinsky2018CausalStructureLearningfromMultivariateTimeSeries}.
However, these methods do not address the problem of inferring causal links between rare events and dynamical systems, across sample trajectories on which the rare event can often occur at different times.


\subsection{Extreme Value Theory and Analysis of Rare Events} 

Extreme value theory characterizes dependences between random variables that exist only when a low-probability event occurs, e.g., rare meteorological events, or financial crises 
\cite{Engelke2022StructureLearningForExtremalTreeModels, Asadi2018OptimalRegionalizationOfExtremeValueDistributionForFloodEstimation}.
Most closely related to our work are \cite{gnecco2021causal}, which studies causal links between heavy-tailed random variables, and \cite{Jana2021CausalAnalysisatExtremeQuantileswithApplicationtoLondonTrafficFlowData}, which explores causal relationships between characteristics of London bicycle lanes, e.g., density, length, and collision rate, and abnormal congestion. However, \cite{gnecco2021causal} imposes restrictive assumptions, such as linear models, while the discussion in \cite{Jana2021CausalAnalysisatExtremeQuantileswithApplicationtoLondonTrafficFlowData} on accidents' occurrences is restricted to empirical studies. In contrast, our proposed algorithm returns a nonparametric conditional independence test statistic that is capable of characterizing relationships between a general dynamical system, and the onset of a rare event.



\subsection{Traffic Network Analysis}

Traffic network theory aims to mathematically describe and control traffic flow in urban networks of roads, bridges, and highways \cite{BaillonCominetti2008MarkovianTrafficEquilibrium, Krichene2014OntheConvergenceofNoRegretLearninginSelfishRouting, 
Ahipasaoglu2019DistributionallyRobustMarkovianTrafficEquilibrium}. Recent literature has proposed the design of tolling mechanisms that drive a traffic network to the socially optimal steady state 
\cite{Maheshwari2022InducingSocialOptimalityinGamesviaAdaptiveIncentiveDesign, Como2022DistributedDynamicPricingofMultiscaleTransportationNetworks}.
However, these methods do not model or predict the occurrence of sudden yet consequential events, such as extreme weather events, car accidents, and other causes of unexpected congestion. 
In contrast, our paper uses the occurrence of rare but consequential car accidents in traffic networks as a running example, to illustrate the applicability of our method on analyzing causal links between dynamical systems and associated rare events.

\section{PRELIMINARIES}
\label{sec: Preliminaries}

Consider a stochastic, discrete-time dynamical system with state variable $X_t \in \R^n$, event variable $A_t \in \{0, 1\}$ with $\Prob(A_t = 1) \in [\pmin, \pmax]$ for some $\pmin, \pmax \in (0, 1)$ for all $t$, with $\pmin < \pmax$, and dynamics $X_{t+1} = f(X_t, A_t, W_t)$ for each $t \geq 0$, where $W_t \in \R^w$ denotes i.i.d. noise, and $f: \R^n \times \{0, 1\} \times \R^w \ra \R^n$ denotes the nonlinear dynamics of the system state. Let $T$ denote the time at which the rare event first occurs, and, with a slight abuse of notation, let $A_{1:t} = 0$ denote the event that $A_1 = \cdots = A_t = 0$. Moreover, we assume that the first occurrence of the rare event is governed by a time-invariant probability distribution, i.e.,:
\begin{align} \label{Eqn: Assumption 1}
    &\Prob(A_{t+1} = 1| X_t \preceq x, A_{1:t} = 0) \\ \nonumber
    = \hspace{0.5mm} &\Prob(A_{t' + 1} = 1| X_{t'} \preceq x, A_{1:t'} = 0), \hspace{5mm} \forall \hspace{0.5mm} t, t' \geq 0,
\end{align}
where, for each $x, y \in \R^n$, the notation $x \preceq y$ represents $x_i \leq y_i$ for each $i \in [n] := \{1, \cdots, n\}$, and for each $x \in \R^n$, there exists some constant ratio $\alpha(x) > 0$ such that:
\begin{align} \label{Eqn: Assumption 2, alpha(x)}
    &\Prob(X_{t-1} \preceq x | A_t = 1, A_{1:t-1} = 0) \\ \nonumber
    = \hspace{0.5mm} &\alpha(x) \cdot \Prob(X_{t-1} \preceq x | A_{1:t-1} = 0).
\end{align}
In words, we assume that the flow distribution is related to the first occurrence of the rare event in a time-invariant manner. Given this setup, we restate \textbf{Q}, first defined in the introduction, as the following hypothesis testing problem:


The binary hypothesis test, with null hypothesis $H_0$ as below, is a mathematically rigorous characterization of \textbf{Q}.

\begin{definition} \label{Def: Binary Hypothesis Test}
Let $H_0$ be the null hypothesis given by:
\begin{align*}
    H_0: \hspace{5mm} &\Prob(A_{t+1} = 1 | X_t \preceq x, A_{1:t} = 0) \\
    = \hspace{0.5mm} &\Prob(A_{t+1} = 1 | A_{1:t} = 0), \hspace{5mm} \forall \hspace{0.5mm} x \in \R, \\
    H_1: \hspace{5mm} &\Prob(A_{t+1} = 1 | X_t \preceq x, A_{1:t} = 0) \\
    \ne \hspace{0.5mm} &\Prob(A_{t+1} = 1 | A_{1:t} = 0), \hspace{5mm} \forall \hspace{0.5mm} x \in \R.
\end{align*}
\end{definition}

In words, $H_0$ holds if and only if the first occurrence of the rare event transpires independently of the system state at that time. For convenience, we define the left and right hand sides of $H_0$ by:
\begin{align} \label{Eqn: Rare event distribution, 1}
    \accprob_1(x) &:= \Prob(A_{t+1} = 1 | X_t \preceq x, A_{1:t} = 0), \\ \label{Eqn: Rare event distribution, 2}
    \accprob_2 &:= \Prob(A_{t+1} = 1 | A_{1:t} = 0).
\end{align}

\textbf{Running Example}: Consider a parallel link traffic network of $R$ links that connect a single source and a single destination. Let $X_{t,i} \in \R$ denote the traffic flow on every link $i \in [R] := \{1, \cdots, R\}$ at time $t$, and define $X_t := (X_{t,1}, \cdots, X_{t, r}) \in \R^R$. (In general, one can define $X_{t, i} \in \R^d$ to encapsulate other observed quantities relevant to link $i$ at time $t$, e.g., vehicle speed and pavement quality). The event variable $A_t = 1$ corresponds to the occurrence of an accident in the network at time $t$.

In this context, Definition \ref{Def: Binary Hypothesis Test} corresponds to checking whether the first occurrence of an accident on the $R$-link network at time $t$ is affected by the flow level at time $t-1$. 
This is of interest to traffic authorities, since costly accidents become more likely at certain levels of traffic flow $X_t$, then the flow should be monitored to decrease the chance that such accidents occur. Flow management can be applied by dynamically tolling the links, as in \cite{Maheshwari2022DynamicTollingforInducingSociallyOptimalTrafficLoads}. As accidents are relatively rare in most traffic datasets, it can be difficult to construct accurate estimates of accident probabilities and flows before accidents at any given time $t$. Instead, below, we propose a novel method of data aggregation that allows the use of information on accident occurrences across all times.

Since $X_t$ is a continuous random variable, a direct comparison of \eqref{Eqn: Rare event distribution, 1} and \eqref{Eqn: Rare event distribution, 2}
would necessitate computing \eqref{Eqn: Rare event distribution, 1} for uncountably many values of $x \in \R^n$. Instead, we use the laws of conditional and total probability to reformulate the problem. In the spirit of Bayes' rule, we compare the state distribution immediately before the rare event occurred, instead of the rare event probabilities under different state values. Formally, under either hypothesis, the state distribution immediately before the first accident can be decomposed as the following infinite sum; for each $x \in \R^n$:
\begin{align*}
     \Prob(X_{T-1} \preceq x) &= \sum_{t=1}^\infty \Prob(X_{t-1} \preceq x, T = t) \\
     &= \sum_{t=1}^\infty \Prob(X_{t-1} \preceq x, A_t = 1, A_{1:t-1} = 0) \\
     &= \sum_{t=1}^\infty \Prob(X_{t-1} \preceq x, A_{1:t-1} = 0) \\
     &\hspace{1cm} \cdot \Prob(A_t = 1| X_{t-1} \preceq x,  A_{1:t-1} = 0).
\end{align*}
Intuitively, if $H_0$ were true, then the condition $X_{t-1} \preceq x$ in the term $\Prob(A_t = 1| X_{t-1} \preceq x,  A_{1:t-1} = 0)$ can be dropped. 
A rigorous formulation is given in Proposition \ref{Prop: Equivalent Condition for H0} below.


\begin{proposition} \label{Prop: Equivalent Condition for H0}
The null hypothesis $H_0$ in Definition \ref{Def: Binary Hypothesis Test} holds if and only if, for each $x \in \R^n$:
\begin{align} \label{Eqn: Equivalent condition for H0}
    &\Prob(X_{T-1} \preceq x) \\ \nonumber
    = \hspace{0.5mm} &\sum_{t=1}^\infty \Prob(X_{t-1} \preceq x, A_{1:t-1} = 0) \cdot \Prob(A_t = 1 | A_{1:t-1} = 0).
\end{align}
\end{proposition}

\vspace{1mm}
\begin{proof}
Please see Appendix \ref{sec: App A, Preliminaries} in the ArXiV version of the paper \cite{Chiu2022TowardsDynamicCausalDiscovery}.
\end{proof}

For convenience, we define, for each $t \in \N$ and $x \in \R$:
\begin{align*}
    \flowatacc_1(x) &:= \Prob(X_{T-1} \preceq x), \\
    \beta_t(x) &:= \Prob(X_{t-1} \preceq x, A_{1:t-1} = 0), \\
    \gamma_t &:= \Prob(A_t = 1 | A_{1:t-1} = 0), \\
    \flowatacc_2(x) &:= \sum_{t=1}^\infty \beta_t(x) \cdot \gamma_t \\
    &= \sum_{t=1}^\infty \Prob(X_{t-1} \preceq x, A_{1:t-1} = 0) \\
    &\hspace{1cm} \cdot \Prob(A_t = 1 | A_{1:t-1} = 0) \\
    &= \sum_{t=1}^\infty \Prob(X_{t-1} \preceq x, A_{1:t-1} = 0) \\
    &\hspace{1cm} \cdot \frac{\Prob(A_t = 1, A_{1:t-1} = 0)}{\Prob(A_{1:t-1} = 0)} \\
    &= \sum_{t=1}^\infty \Prob(X_{t-1} \preceq x| A_{1:t-1} = 0) \cdot \Prob(T=t).
\end{align*}
Note that $b_1(x), b_2(x) \in [0, 1]$ (in particular, that $b_2(x) \leq 1$ follows by observing that $b_2(x) \leq \sum_{t=1}^\infty \Prob(T = t) = 1$.)

The test statistic that we use to distinguish between the distributions $b_1(x)$ and $b_2(x)$ is the gap: 
\begin{align*}
    \sup_{x \in \R^n} |b_1(x) - b_2(x)|
\end{align*}
Intuitively, a large gap would indicate a higher likelihood that a component-wise larger or smaller state would change the probability of an event occurring. We formalize this notion in Algorithm \ref{Alg: Hypothesis Testing}, and provide finite sample guarantees for empirical estimates of $b_1(x)$ and $b_2(x)$ that can be constructed efficiently from data and used to compute the test statistic.

\textbf{Running Example}: In the traffic network example, $b_1(x)$ corresponds to the probability that $X_{T-1}$, the network flows before the first accident, is component-wise less than or equal to $x$. Meanwhile, $b_2(x)$ describes the weighted average of traffic flows at each time $t$, conditioned on the first accident occurring after $t$ (i.e., no accident occurs before), with the distribution of the first accident time $T$ as weights. Section \ref{sec: Methods} describes sample-efficient methods for constructing empirical estimates of $b_1(x)$ and $b_2(x)$ from a dataset of independent traffic flows.

\section{METHODS}
\label{sec: Methods}



\subsection{Main Algorithm}

We present Algorithm \ref{Alg: Hypothesis Testing}, which solves the hypothesis testing problem in Definition \ref{Def: Binary Hypothesis Test} from a dataset of $N$ independent trajectories, by constructing and comparing finite-sample empirical cumulative distribution functions (CDFs) $\hat b_1^N(x)$ and $\hat b_2^N(x)$ for the expressions $b_1(x)$ and $b_2(x)$, respectively, and verifying whether or not \eqref{Eqn: Equivalent condition for H0} holds (in accordance with Proposition \ref{Prop: Equivalent Condition for H0}). 

\paragraph{Note on the baseline method} The common baseline method for resolving the problem in Definition \ref{Def: Binary Hypothesis Test} is to fix $t \geq 1$, and compare the CDF values $\Prob(X_{t-1} \preceq x|T = t)$ and $\Prob(X_{t-1} \preceq x)$, for each $x \in \R^n$ at the fixed $t$. This is effectively a \say{static variant} of Algorithm \ref{Alg: Hypothesis Testing} that only utilizes dynamical state values immediately before accidents that occur at time $t$. 
It is generally difficult to estimate $\Prob(X_{t-1} \preceq x | T =t)$ from data, since $\Prob(T = t)$ can be very small for any given $t$. Our algorithm (Algorithm \ref{Alg: Hypothesis Testing}) instead aggregates data \emph{across} times when the rare event has occurred,  allowing the event to be represented with higher probability.




\begin{algorithm} 
    {
    \small
    \SetAlgoLined
    \KwData{Dataset of system state and rare event variables: $\{(X_t^i, A_t^i): t \geq 0, i \in [N]\}$}
    
    \KwResult{Distribution gap: $\sup_{x \in \R} |\hat b_1^N(x) - \hat b_2^N(x)|$}
    
    $\hat T^i \gets$ Realization of $T$ for data trajectory $i$, $\forall \hspace{0.5mm} i \in [N]$.
    
    $\hat b_1^N(x) \gets \frac{1}{N} \sum_{i=1}^N \textbf{1}\{X_{\hat T^i - 1} \preceq x\}$.
    
    $\hat \beta_t^N(x) \gets \frac{1}{N} \sum_{i=1}^N \textbf{1}\{X_t^i \preceq x, A_{1:t-1}^i = 0 \}$.
    
    $\hat \gamma_t^N \gets \begin{cases}
        \frac{\sum_{i=1}^N \textbf{1}\{A_{1:t-1}^i = 0, A_t^i = 1\}}{\sum_{i=1}^N \textbf{1}\{A_{1:t-1}^i = 0\}}, \hspace{5mm} &\text{if } \sum_{i=1}^N \textbf{1}\{A_{1:t-1}^i = 0\} > 0, \\
        0, \hspace{5mm} &\text{else}.
    \end{cases}$.
    
    $\hat b_2^N(x) \gets \sum_{t=1}^\infty \hat \beta_t^N(x) \cdot \hat \gamma_t^N $. \label{Eqn: b 2 hat N, in Algorithm}
    
    \Return{$\sup_{x \in \R}|\hat b_1^N(x) - \hat b_2^N(x)|$.}
     }
     \caption{Hypothesis Testing with Reorganized Dataset.}
     \label{Alg: Hypothesis Testing}
\end{algorithm}


\subsection{Theoretical Guarantees}

Theorem \ref{Thm: Exponential Convergence to Consistency Against all Alternatives} below illustrates that, if $H_0$ holds, then as the number of sample trajectories $N$ approaches infinity, the empirical distributions of \eqref{Eqn: Flow at rare event, distribution, 1} and \eqref{Eqn: Flow at rare event, distribution, 2}, as constructed in Algorithm \ref{Alg: Hypothesis Testing}, converge at an exponential rate to their true values. In other words, if $H_0$ holds, then for any fixed significance level $\alpha$, Algorithm \ref{Alg: Hypothesis Testing} will require a dataset of size no greater than $O(\ln(1/\alpha))$ to reject $H_1$. This establishes a finite sample bound that controls the error of the statistical independence test corresponding to the test statistic presented in Algorithm \ref{Alg: Hypothesis Testing}. The proof follows by carefully applying concentration bounds for light-tailed random variables, and invoking the Dvoretsky-Kiefer-Wolfowitz (DKW) inequality \cite{Dvoretzky1956AsymptoticMinimaxCharacteroftheSampleDistributionFunction}, which prescribes explicit convergence rates for empirical CDFs to the true CDF.

\begin{theorem}(\textbf{\textit{Exponential Convergence to Consistency Against all Alternatives}}) \label{Thm: Exponential Convergence to Consistency Against all Alternatives}
Suppose the null hypothesis $H_0$ holds, i.e., $\flowatacc_1(x) = \flowatacc_2(x)$. 
\begin{enumerate}
    \item If $n = 1$, i.e., $X_t \in \R$ for each $t \geq 0$, then for each $\epsilon > 0$, there exist continuous, positive functions $C_1(\epsilon), C_2(\epsilon) > 0$ such that:
    \begin{align*}
        &\Prob\left(  
        \sup_{x \in \R^n} \Big\{ \big|\hat \flowatacc_1^N(x) - \hat \flowatacc_2^N(x) \big| \Big\} > \epsilon \right) \\
        \leq \hspace{0.5mm} &C_1(\epsilon) \cdot e^{-N \cdot C_2(\epsilon)}.
    \end{align*}

    \item If $n > 1$, then there exist continuous, positive functions $C_3(\epsilon), C_4(\epsilon) > 0$ such that:
    \begin{align*}
        &\Prob\left(  
        \sup_{x \in \R^n} \Big\{ \big|\hat \flowatacc_1^N(x) - \hat \flowatacc_2^N(x) \big| \Big\} > \epsilon \right) \\
        \leq \hspace{0.5mm} &\Big[ C_3(\epsilon) (N+1) n + C_4(\epsilon) \Big] \cdot e^{-N \cdot C_5(\epsilon)}.
    \end{align*}
    For sufficiently large $N$, the factor $N+1$ can be replaced by the constant $2$.
\end{enumerate}
\end{theorem}

\begin{proof}
Please see Appendix \ref{sec: App B, Methods} in the ArXiV version of the paper \cite{Chiu2022TowardsDynamicCausalDiscovery}.
\end{proof}

\begin{remark}
If $H_0$ does not hold, i.e., $\delta := \sup_{x \in \R^n}|b_1(x) - b_2(x)| > 0$, then the same logical arguments used to establish Theorem \ref{Thm: Exponential Convergence to Consistency Against all Alternatives} can be employed to show that (for the $n = 1$ case), for each $\epsilon > 0$:
\begin{align*}
    &\Prob\left(  
    \sup_{x \in \R^n} \Big\{ \big|\hat \flowatacc_1^N(x) - \hat \flowatacc_2^N(x) \big| \Big\} \in (\delta - \epsilon, \delta + \epsilon) \right) \\
    \leq \hspace{0.5mm} &C_1(\epsilon) \cdot e^{-N \cdot C_2(\epsilon)},
\end{align*}
where $C_1(\epsilon)$, $C_2(\epsilon) > 0$ are the same continuous, positive functions given above. That is, as $N \ra \infty$, the gap between $\flowatacc_1^N(x)$ and $\flowatacc_2^N(x)$ approaches $\delta$ exponentially. The $n > 1$ case follows analogously from the multivariate version of the Dvoretsky-Kiefer-Wolfowitz (DKW) inequality \cite{Naaman2021OnTheTightConstantintheMultivariateDKWInequality}.
\end{remark}

\section{RESULTS}
\label{sec: Results}

Here, we illustrate the numerical performance of our proposed method on simulated and real-world traffic data, and its efficacy over baseline aggregation methods of concatenating data points along a single, fixed time $t$. We note that in the experiments on the real-world dataset collected from the Caltrans PeMS system, the data collected is time-ordered. Code containing the datasets and experiments is publicly available at the following link: \url{https://github.com/kkulk/L4DC2023-Causality}.


\subsection{Simulated Data}
\label{subsec: Simulated Data}

In our first set of experiments, we construct synthetic data for single- and multi-link traffic networks. For the single-link network, we use the following dynamics. For each $t \in [\Thorizon]$:
\begin{align*}
    x[t+1] &= (1-\mu(A[t]) ) \cdot x(t) + \mu(A[t]) \cdot u[t] + w[t], \\
    A[t+1] &\sim \mathcal{P}(x(t))
\end{align*}
where $x(t) \in \R$ denotes the traffic flow at time $t$, $A[t] \in \{0, 1\}$ is the Boolean random variable that indicates whether or not an accident has occurred at time $t$, $\mu(A[t]) > 0$ describes the fraction of traffic flow departing the link, $u[t] \in \R$ denotes the total input traffic flow, $w[t] \in \R$ is a zero-mean noise term, and $\Thorizon$ is the finite time horizon. Here, we set $\Thorizon = 500$, $\mu(0) = 0.3$, $\mu(1) = 0.2$, $u(t) = 100$ for each $t \in [\Thorizon]$, and draw $w(t)$ i.i.d. from the continuous uniform distribution on $(-10, 10)$. We create datasets corresponding to the null and alternative hypotheses. For the null hypothesis, we fix the distribution of $x(t)$ to be Bernoulli($0.01$), regardless of the value of $x(t)$. This simulates a scenario where the likelihood of an accident occurring has no dependence on traffic flow. For the alternative hypothesis, we set the distribution of $x(t)$ to be Bernoulli($0.01$) when $x(t) < 109$ and Bernoulli($0.10$) when $x(t) \geq 109$. This represents a scenario where higher traffic loads increase the likelihood that an accident occurs.

\begin{figure*}[ht]
    \centering
    \includegraphics[scale=0.2]{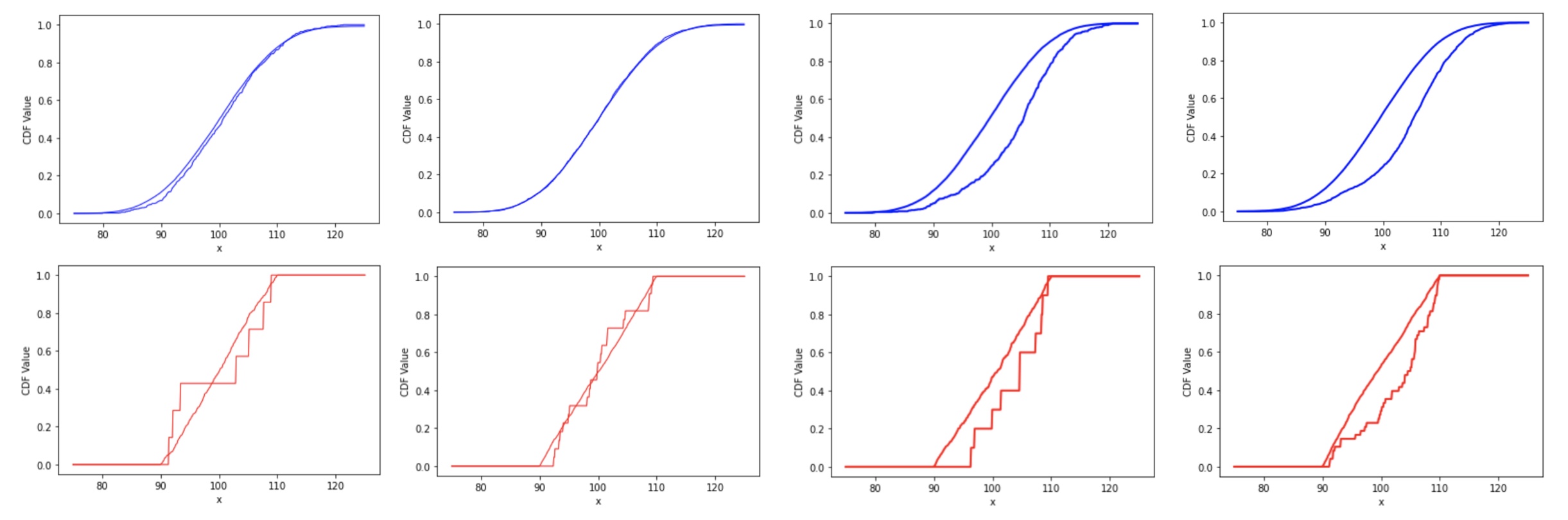}
    \caption{(Top) From left to right, $b_1(x)$ and $b_2(x)$ vs. $x$ plots for $(H_0, N = 500)$, $(H_0, N = 2000)$, $(H_1, N = 500)$, and $(H_1, N = 2000)$.
    (Bottom) From left to right, empirical CDFs for $X_{t-1}|T=t$ and $X_{t-1}$ with $t = 1$, in the same order of hypothesis and $N$ values.}
    \label{fig: CDF_vs_x___single_link_combined}
\end{figure*}

\begin{figure}
    \centering
    \includegraphics[scale=0.14]{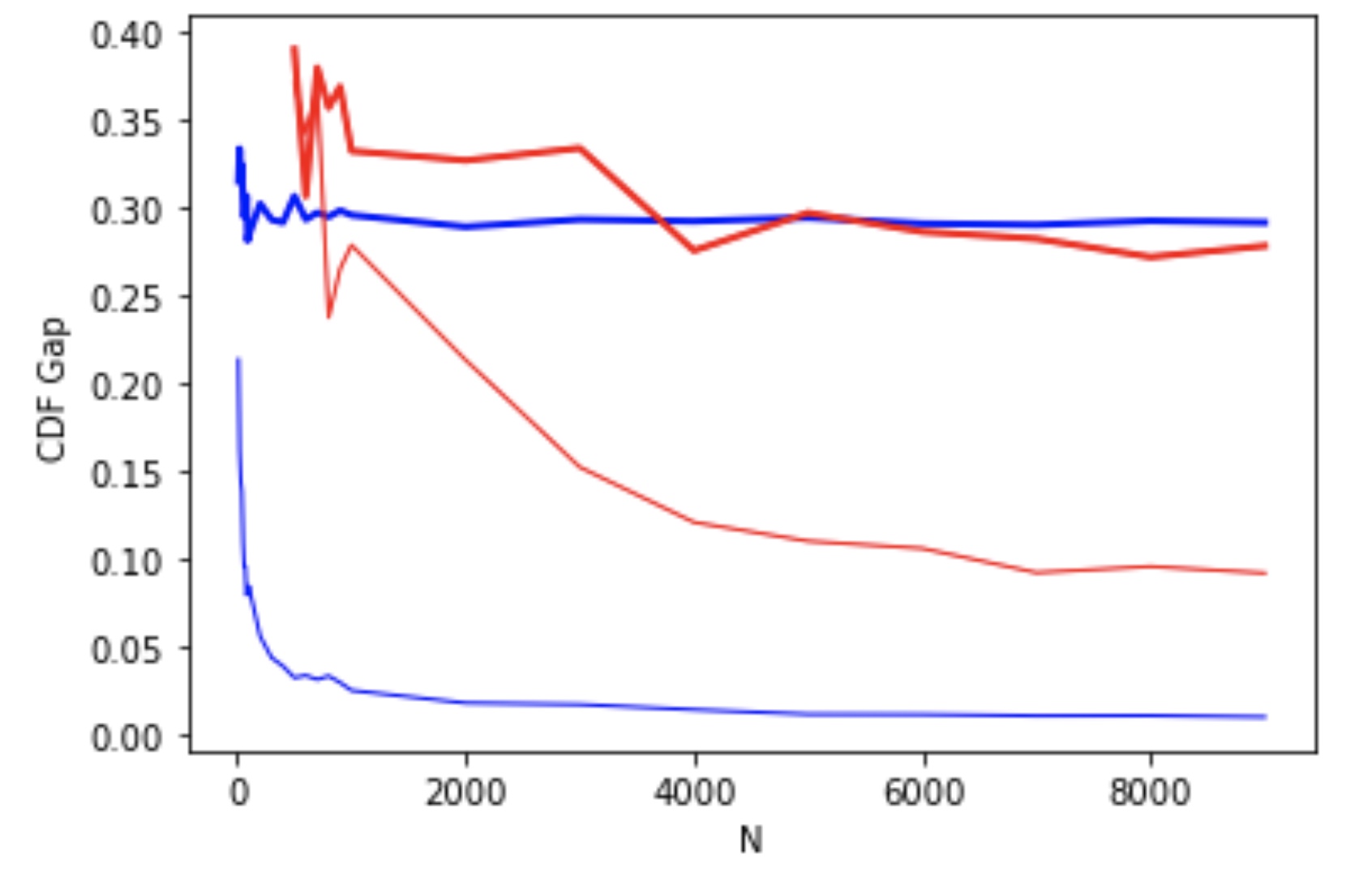}
    \caption{
    CDF gap between vs. $N$. 
    Red and blue correspond to the baseline and our method, respectively, while thick and thin lines correspond to the null and alternative hypotheses, respectively. 
    Our approach (thin blue curve) correctly identifies the null hypothesis dataset with a relatively small number of samples, while the baseline aggregation method fails to do so (thin red curve).}
    \label{fig: CDF_Error_vs_N___single_link}
\end{figure}

To contrast the performance of our algorithm with the baseline, we compute the following quantities from datasets of independent trajectories corresponding to $H_0$ and $H_1$, in accordance with Proposition \ref{Prop: Equivalent Condition for H0} and Theorem \ref{Thm: Exponential Convergence to Consistency Against all Alternatives}:
\begin{itemize}
    \item \textbf{For our method}---We compute the empirical estimates $\hat b_1^N(x)$ and $\hat b_2^N(x)$ of the functions $b_1(x)$ and $b_2(x)$ as functions of $x$ (Figure \ref{fig: CDF_vs_x___single_link_combined}), and the maximum CDF gap $\sup_{x \in \R^n}|\hat b_1^N(x) - \hat b_2^N(x)|$ as functions of $N$ (Figure \ref{fig: CDF_Error_vs_N___single_link}). 
    
    \item \textbf{For the baseline method}---We compute the empirical estimates of the CDFs of $X_{t-1}|T=t$ and $X_{t-1}$, with $t$ fixed at 1, as functions of $x$ (Figure \ref{fig: CDF_vs_x___single_link_combined}), and the corresponding maximum CDF gap as functions of $N$ (Figure \ref{fig: CDF_Error_vs_N___single_link}). Note that for $N < 500$, it is difficult to obtain the CDF of $X_{t-1}|T=t$, due to rarity of the event at any given time.
\end{itemize}
Figures \ref{fig: CDF_vs_x___single_link_combined} and \ref{fig: CDF_Error_vs_N___single_link} show that, compared to the baseline, our approach distinguishes between the null and alternative hypotheses from a far smaller dataset. This illustrates that our method, compared to the baseline, distinguishes the dependence between the occurrence of a rare event and the state values immediately preceding the event more efficiently. 

Appendix \ref{sec: App C, Experiment Results} contains further empirical results on synthetic datasets for multi-link networks.

\subsection{Caltrans PeMS Dataset}

We also demonstrate the efficacy of Algorithm \ref{Alg: Hypothesis Testing} on real traffic flow and incident data collected from the publicly available Caltrans Performance Measurement System (PeMS) dataset \cite{varaiya2001freeway}. PeMS uses loop detectors placed on freeways to collect flow, speed, and other traffic condition information, and overlays this with incident reports. We consider daily traffic flow data (\# vehicles / time) collected from January to August 2022 from 6 A.M. to 2 P.M., at 5-minute intervals, on various bridges in the San Francisco Bay Area: San Mateo-Hayward, San Francisco–Oakland Bay, and Richmond-San Rafael. That is, we consider single link networks connecting a source and destination with the continuous variables $X_t \in \mathbb{R}_{+}$ corresponding to average flows on the link. 
Correspondingly, we use incident data collected on these bridges by PeMS in the same time interval from the California Highway Patrol (CHP).

\paragraph{Data Collection}
We treat each day 
as an independent trajectory of the traffic flows on every bridge. The PeMS dataset contains flows collected from dual loop detectors placed along the bridges. For each time between 6 A.M. and 2 P.M., we average the flow data recorded by loop detectors on each bridge to obtain the state variable $X_t$ for time $t$. Mathematically, we define $X_t := \frac{1}{|I|} \sum_{i=1}^{|I|} X_t^i$, where $I$ denotes the set of loop detectors on a single link, and $X_t^i$ denotes the flow measured by detector $i\in I$ at time $t$. 
We exclude from our analysis any trajectory on which there was no incident for the entire day, since such trajectories do not contain data relevant to our problem of interest.

\paragraph{Results}
In Table 1, we enumerate the sample size $N$ and test statistic $\sup_{x\in \R} |\hat{b}_1^N(x) - \hat{b}_2^N(x)|$ for the six traffic links (three bridges, each with two directions of traffic flow). Note the substantial difference in the CDF gap (of nearly 0.178) for the Richmond-San Rafael Bridge, East, compared to all other links, indicating that the flows on this link are particularly causally linked to the first time of incident formation. Further, the San Francisco-Oakland Bay Bridge, East, also has a higher CDF gap (0.081) relative to the West direction, and relative to the other bridges. These gaps are visible in the CDF plots in Figures 3(e) and 3(c), respectively. 

\begin{table}[!h]
\label{table: bridges-cdf} 
\begin{center}
{
\begin{tabular}{|c| c| c|}
    \hline 
    \textbf{Link} & \textbf{$N$} & \textbf{$\sup\limits_{x \in \R} |\hat{b}_1^N(x) - \hat{b}_2^N(x)|$} \\
    \hline 
    San Mateo-Hayward & 85 & 0.053  \\
    Bridge, East (SR92-E) & & \\
    \hline 
    San Mateo-Hayward & 116 & 0.039 \\ 
    Bridge, West (SR92-W) & & \\
    \hline 
     San Francisco–Oakland & 116 & 0.081 \\ 
     Bay Bridge, East (I80-E) & & \\
     \hline
      San Francisco–Oakland & 112 & 0.042 \\ 
      Bay Bridge, West (I80-W) & & \\
      \hline
    Richmond-San Rafael & 45 & 0.178 \\ 
    Bridge, East (I580-E) & & \\
    \hline 
    Richmond-San Rafael & 94 & 0.048 \\
    Bridge, West (I580-W) & & \\ 
    \hline 
\end{tabular}
}
\end{center}
\caption{CDF gap, $\sup_{x \in \R^n} |\hat{b}_1^N(x) - \hat{b}_2^N(x)|$ for the six links in the San Francisco Bay Area.}
\end{table}



\begin{figure*} 
    \centering
  \subfloat[San Mateo-Hayward Bridge, East (SR92-E)\label{subfig:SR92-E}]{%
       \includegraphics[width=0.5\linewidth]{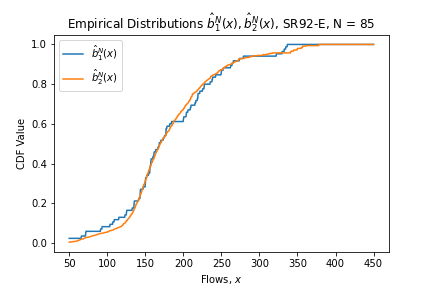}}
    \hfill
  \subfloat[San Mateo-Hayward Bridge, West (SR92-W)\label{subfig:SR92-W}]{%
        \includegraphics[width=0.5\linewidth]{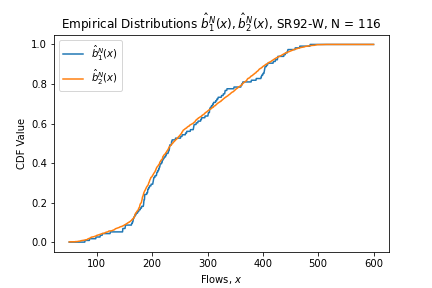}}
    \\
  \subfloat[San Francisco–Oakland Bay Bridge, East (I80-E)\label{subfig:I80-E}]{%
        \includegraphics[width=0.5\linewidth]{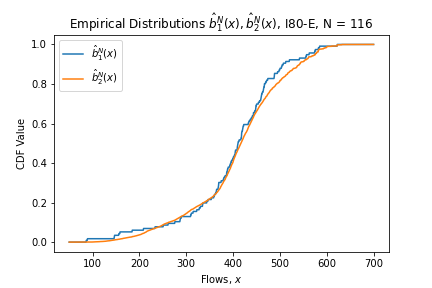}}
    \hfill
  \subfloat[San Francisco–Oakland Bay Bridge, West (I80-W)\label{subfig:I80-W}]{%
        \includegraphics[width=0.5\linewidth]{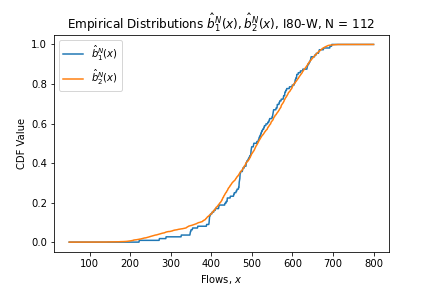}}
    \\
  \subfloat[Richmond-San Rafael Bridge, East (I580-E)\label{subfig:I580-E}]{%
        \includegraphics[width=0.5\linewidth]{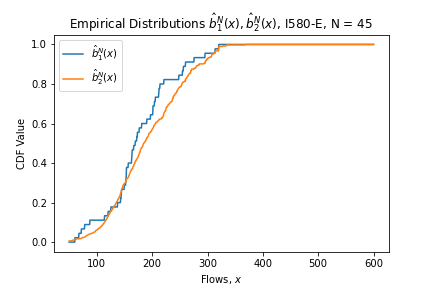}}
    \hfill
  \subfloat[Richmond-San Rafael Bridge, West (I580-W)\label{subfig:I580-W}]{%
        \includegraphics[width=0.5\linewidth]{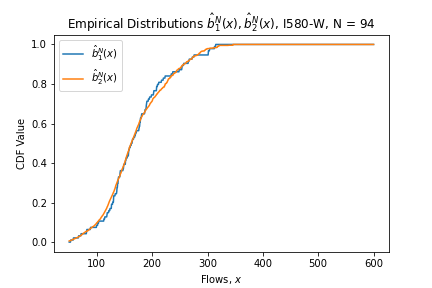}}
  \caption{Empirical CDFs $\hat{b}_1^N(x)$ and $\hat{b}_2^N(x)$ for six bridges in the San Francisco Bay Area.}
  \label{fig:rsrw} 
\end{figure*}


\section{CONCLUSION AND FUTURE WORK}
\label{sec: Conclusion and Future Work}

We present a novel method for identifying causal links between the state evolution of a dynamical system and the onset of an associated rare event. 
Crucially, we leverage the time-invariance to reorganize data in a manner that better represents occurrences of the rare event. We then formulate a non-parametric statistical independence test to infer causal dependencies between the dynamical states and the rare event. Empirical results on simulated and real-world time-series data indicate that our method outperforms a baseline approach that conducts independence tests only on a single time slice of the original rare events dataset.

As future work, the causal discovery algorithm presented here may be used to more effectively control the evolution of a dynamical system associated with a rare but consequential event. By establishing causal links between the dynamical state and the rare event, control strategies can be redesigned to maneuver the state away from regions of the state space where the event occurs more frequently. Important engineering applications include incentive design and flow control methods in the network traffic systems literature, such as dynamic tolling and rerouting. 
Finally,
we will present more extensive empirical analysis of both the baseline method and our method across different applications.

\printbibliography

\appendix

Please use the following link to access an ArXiV version \cite{Chiu2022TowardsDynamicCausalDiscovery} with the appendix (\url{https://arxiv.org/pdf/2211.16596.pdf}). The authors will ensure this link stays active.

\newpage

The following supplementary material includes the appendix, which contains proofs and figures omitted in the main paper due to space limitations.

\subsection{Preliminaries}
\label{sec: App A, Preliminaries}


\begin{proof}(\textbf{Proof of Proposition \ref{Prop: Equivalent Condition for H0}})
Fix $x \in \R^n$ arbitrarily. By Bayes' rule:
\begin{align*}
    &\Prob(A_t = 1 | X_{t-1} \leq x, A_{1:t-1} = 0) \\
    = \hspace{0.5mm} &\Prob(A_t = 1 | A_{1:t-1} = 0) \\
    &\hspace{1cm} \cdot \frac{\Prob(X_{t-1} \preceq x | A_t = 1, A_{1:t-1} = 0)}{\Prob(X_{t-1} \preceq x | A_{1:t-1} = 0)} \\
    = \hspace{0.5mm} &\Prob(A_t = 1 | A_{1:t-1} = 0) \cdot \alpha(x),
\end{align*}
where $\alpha(x)$ is as defined in \eqref{Eqn: Assumption 2, alpha(x)}. Thus, $\Prob(A_t = 1| A_{t-1} = 0)$ is time-invariant, i.e., holds the same value for each $t \in [T]$. Next, by invoking Bayes' rule again, we have:
\begin{align} \label{Eqn: Flow at rare event, distribution, 1}
    &\Prob(X_{T-1} \preceq x) \\ \nonumber
    = \hspace{0.5mm} &\sum_{t=1}^\infty \Prob(X_{t-1} \preceq x| T = t) \cdot \Prob(T = t) \\ \nonumber
    = \hspace{0.5mm} &\sum_{t=1}^\infty \Prob(X_{t-1} \preceq x, T = t) \\ \nonumber
    = \hspace{0.5mm} &\sum_{t=1}^\infty \Prob(X_{t-1} \preceq x, A_{1:t-1} = 0, A_t = 1) \\ \nonumber
    = \hspace{0.5mm} &\sum_{t=1}^\infty \Prob(A_t = 1 | X_t \preceq x, A_{1:t-1} = 0) \\ \nonumber
    &\hspace{1cm} \cdot \Prob(X_{t-1} \preceq x, A_{1:t-1} = 0) \\ \nonumber
    = \hspace{0.5mm} &a_1(x) \cdot \sum_{t=1}^\infty \Prob(X_{t-1} \preceq x, A_{1:t-1} = 0).
\end{align}
and:
\begin{align} \label{Eqn: Flow at rare event, distribution, 2}
    &\sum_{t=1}^\infty \Prob(X_{t-1} \preceq x, A_{1:t-1} = 0) \\ \nonumber
    &\hspace{1cm} \cdot \Prob(A_t = 1 | A_{1:t-1} = 0) \\ \nonumber
    = \hspace{0.5mm} &a_2 \cdot \sum_{t=1}^\infty \Prob(X_{t-1} \preceq x, A_{1:t-1} = 0). 
\end{align}
Thus, the null hypothesis $H_0$ in Definition \ref{Def: Binary Hypothesis Test} holds if and only if \eqref{Eqn: Flow at rare event, distribution, 1} and \eqref{Eqn: Flow at rare event, distribution, 2} are equal, as claimed.
\end{proof}

\subsection{Methods}
\label{sec: App B, Methods}

\begin{proof}(\textbf{Proof of Theorem \ref{Thm: Exponential Convergence to Consistency Against all Alternatives}})
Fix $\epsilon > 0$, and take: 
\begin{align*}
    \Tcutoff := \left\lceil \frac{1}{\ln(1 - \pmin)} \ln\left( \frac{\epsilon\pmin^2}{16 \pmax} \right) \right\rceil.
\end{align*}
First, to show that $\hat b_1^N(x) \ra b_1(x)$ at an exponential rate in $N$, we invoke the Dvoretsky-Kiefer-Wolfowitz inequality:
\begin{align*}
    \Prob\left( \sup_{x \in \R^n} \big|\hat b_1^N(x) - b_1(x) \big| > \frac{1}{2}\epsilon \right) \leq 2 \cdot e^{-\frac{1}{2} N \epsilon^2}
\end{align*}
Next, to show that $\hat b_2^N(x) \ra b_2(x)$ at an exponential rate in $N$, we have, via the triangle inequality:
\begin{align*}
    &\sup_{x \in \R^n} \Bigg| \hat \flowatacc_2^N(x) - \flowatacc_2(x) \Bigg| \\
    = \hspace{0.5mm} &\sup_{x \in \R^n} \left| \sum_{t=1}^\infty \Big[ \hat \beta_t^N(x) \hat \gamma_t^N - \beta_t(x) \gamma_t \Big] \right| \\
    = \hspace{0.5mm} &\sum_{t=1}^{\Tcutoff} \Big[ \sup_{x \in \R^n} \Big\{|\hat \beta_t^N(x) - \beta_t(x)|\Big\} \hat \gamma_t^N \\
    &\hspace{2cm} + \sup_{x \in \R^n} \Big\{|\hat \gamma_t^N - \gamma_t|\Big\} \beta_t(x) \Big] \\
    &\hspace{1cm} + \sup_{x \in \R^n} \Bigg\{ \sum_{t=\Tcutoff+1}^\infty \Big[ |\hat \beta_t^N(x) \hat \gamma_t^N| + |\beta_t(x) \gamma_t| \Big] \Bigg\} \\
    \leq \hspace{0.5mm} &\sum_{t=1}^{\Tcutoff} \sup_{x \in \R^n} \Big\{ |\hat \beta_t^N(x) - \beta_t(x)| \Big\} \\
    &\hspace{1cm} + \sum_{t=1}^{\Tcutoff} \sup_{x \in \R^n} \Big\{ |\hat \gamma_t^N - \gamma_t| \Big\} \cdot \Prob(A_{1:t-1} = 0) \\
    &\hspace{1cm} + \frac{1}{N} \sum_{i=1}^N \sum_{t = \Tcutoff+1}^\infty \textbf{1}\{\hat T^i = t \} + \sum_{t = \Tcutoff+1}^\infty \Prob(T = t),
\end{align*}
where the third and fourth term in the final expression follow by observing that, for any $x \in \R$, by definition of the quantities $\hat \beta_t^N(x)$, $\hat \gamma_t^N$, $\beta_t(x)$, and $\gamma_t$:
\begin{align*}
    &|\hat \beta_t^N(x) \hat \gamma_t^N| \\
    = \hspace{0.5mm} &\frac{1}{N} \sum_{i=1}^N \textbf{1}\{X_t^i \preceq x, A_{1:t-1}^i = 0\} \\
    &\hspace{1cm} \cdot \frac{\sum_{i=1}^N \textbf{1}\{A_{1:t-1}^i = 0, A_t^i = 1\}}{\sum_{i=1}^N \textbf{1}\{A_{1:t-1}^i = 0\}} \\
    \leq \hspace{0.5mm} &\frac{1}{N} \sum_{i=1}^N \textbf{1}\{A_{1:t-1}^i = 0\} \\
    &\hspace{1cm} \cdot \frac{\sum_{i=1}^N \textbf{1}\{A_{1:t-1}^i = 0, A_t^i = 1\}}{\sum_{i=1}^N \textbf{1}\{A_{1:t-1}^i = 0\}} \\
    = \hspace{0.5mm} &\frac{1}{N} \sum_{i=1}^N \textbf{1}\{A_{1:t-1}^i = 0, A_t^i = 1\} \\
    = \hspace{0.5mm} &\frac{1}{N} \sum_{i=1}^N \textbf{1}\{\hat T^i = t \}
\end{align*}
and similarly:
\begin{align*}
    &|\beta_t(x) \gamma(t)| \\
    = \hspace{0.5mm} &\Prob(X_{t-1} \preceq x, A_{1:t-1} = 0) \cdot \Prob(A_t = 1 | A_{1:t-1} = 0) \\
    \leq \hspace{0.5mm} &\Prob(A_{1:t-1} = 0) \cdot \Prob(A_t = 1 | A_{1:t-1} = 0) \\
    \leq \hspace{0.5mm} &\Prob(A_t = 1, A_{1:t-1} = 0) \\
    = \hspace{0.5mm} &\Prob(T = t).
\end{align*}
Below, we upper bound each of the four terms in the final expression above.

\begin{itemize}
    \item First, by the Dvoretsky-Kiefer-Wolfowitz inequality, we have, for each $t \in [\Tcutoff] := \{1, \cdots, \Tcutoff\}$:
    \begin{align*}    
        &\Prob\left( \sum_{t=1}^{\Tcutoff} \sup_{x \in \R^n} \Big\{ |\hat \beta_t^N(x) - \beta_t(x)| \Big\} \geq \frac{1}{8} \epsilon \right) \\
        \leq \hspace{0.5mm} &\Prob\left( \bigcup_{t=1}^{\Tcutoff} \Bigg\{ \sup_{x \in \R^n} \Big\{ |\hat \beta_t^N(x) - \beta_t(x)| \Big\} \geq \frac{1}{8T_c} \epsilon \Bigg\} \right) \\
        \leq \hspace{0.5mm} &\sum_{t=1}^{\Tcutoff} \Prob\left( \sup_{x \in \R^n} \Big\{ |\hat \beta_t^N(x) - \beta_t(x)| \Big\} \geq \frac{1}{8 \Tcutoff} \epsilon \right) \\
        \leq \hspace{0.5mm} &2 \Tcutoff \exp\left( - \frac{\epsilon^2}{32 \Tcutoff^2} \cdot N \right).
    \end{align*}
    
    \item Second, let $N_t \in [N]$ denote the number of trajectories with $A_{1:t-1} = 0$. We first show that, with high probability, $N_t \geq N \cdot \Prob(A_{1:t-1} = 0)^2$. We then show that, under this condition on $N_t$ taking a sufficiently large value, $\hat \gamma_t^N(x) \ra \gamma_t(x)$ exponentially in $N$.
    
    $\hspace{5mm}$ First, the Hoeffding bound for general bounded random variables (\cite{Vershynin2018HighDimensionalProbability} Theorem 2.2.6) gives:
    \begin{align*}
        &\Prob\left( \frac{1}{N} N_t \leq \Prob(A_{1:t-1} = 0)^2 \right) \\
        \leq \hspace{0.5mm} &\Prob\Bigg( \Bigg| \frac{1}{N} N_t - \Prob(A_{1:t-1} = 0) \Bigg| \\
        &\hspace{1cm} \geq \Prob(A_{1:t-1} = 0) - \Prob(A_{1:t-1} = 0)^2 \Bigg) \\
        \leq \hspace{0.5mm} &\exp\left( - 2 \Big[ \Prob(A_{1:t-1} = 0) - \Prob(A_{1:t-1} = 0)^2 \Big]^2 N \right)
    \end{align*}
    Then, if $N_t \geq N \cdot \Prob(A_{1:t-1} = 0)$, we can bound the gap between $\hat \gamma_t^N(x)$ and $\gamma_t(x)$ as follows:
    \begin{align*}
        &\Prob\left( |\hat \gamma_t^N(x) - \gamma_t(x)| > \frac{\epsilon}{8\Tcutoff \cdot \Prob(A_{1:t-1} = 0)} \right) \\
        \leq \hspace{0.5mm} &\exp\Bigg( - 2 \cdot \Prob(A_{1:t-1} = 0)^2 \cdot N \\
        &\hspace{1cm} \cdot \frac{\epsilon^2}{64 \Tcutoff^2 \cdot \Prob(A_{1:t-1} = 0)^2} \Bigg) \\
        \leq \hspace{0.5mm} &\exp\left( - \frac{\epsilon^2}{32 \Tcutoff^2} \cdot N \right).
    \end{align*}
    
    \item Third, to bound  $\hat B_{\Tcutoff}^N := \frac{1}{N} \sum_{i=1}^N \sum_{t=\Tcutoff+1}^\infty \textbf{1}\{\hat T^i = t \}$, define:
    \begin{align*}
        B_{\Tcutoff} &:= \sum_{t = \Tcutoff+1}^\infty \textbf{1}\{T = t\} =  \textbf{1}\{T > T_c\}.
    \end{align*}
    Thus, $B_{\Tcutoff}$ is a Bernoulli random variable with parameter $\Prob(B_{T_c} = 1)$, and expectation upper bounded by:
    \begin{align*}
        \E[B_{T_c}] &= \Prob(B_{T_c} = 1) \leq (1 - p_1)^{T_c}.
    \end{align*}
    By definition of $T_c$, we have $\E[B_{T_c}] \leq \frac{1}{16}\epsilon$. Moreover, since $B_{T_c}$ is a Bernoulli random variable, we have, by the Hoeffding bound for general bounded random variables (\cite{Vershynin2018HighDimensionalProbability} Theorem 2.2.6):
    \begin{align*}
        &\Prob\left( \frac{1}{N} \sum_{i=1}^N \hat B_{\Tcutoff}^N > \frac{1}{8} \epsilon \right) \\
        = \hspace{0.5mm} &\Prob\left( \frac{1}{N} \sum_{i=1}^N \hat B_{\Tcutoff}^N - \E[B_{\Tcutoff}] > \frac{1}{8} \epsilon - \E[B_{\Tcutoff}] \right) \\
        \leq \hspace{0.5mm} &\Prob\left( \frac{1}{N} \sum_{i=1}^N \hat B_{\Tcutoff}^N - \E[B_{\Tcutoff}] > \frac{1}{16} \epsilon \right) \\
        < \hspace{0.5mm} &\exp\left( - \frac{1}{128}N \right).
    \end{align*}
    
    \item Finally, note that by definition of $T_c$:
    \begin{align*}
        \sum_{t=\Tcutoff+1}^\infty \Prob(T = t) &= \Prob(T > T_c) \\
        &\leq (1-p_1)^{T_c} \\
        &< \frac{1}{16} \epsilon.
    \end{align*}
\end{itemize}

For the multivariate version (i.e., $n > 1$), the same proof follows, albeit with the multivariate version of the Dvoretsky-Kiefer-Wolfowitz inequality \cite{Naaman2021OnTheTightConstantintheMultivariateDKWInequality}.
\end{proof}

\subsection{Experiment Results}
\label{sec: App C, Experiment Results}

\subsubsection{Multi-link Traffic Networks}
\label{subsec: App C, Multi-link Traffic Networks}

For the multi-link traffic network, we use the dynamics: (\cite{Maheshwari2022DynamicTollingforInducingSociallyOptimalTrafficLoads})
\begin{align} \label{Eqn: Multi-link Traffic Network}
    &x_i[t+1] \\
    = \hspace{0.5mm} &(1-\mu) \cdot x_i[t] + \mu \cdot \frac{e^{-\beta \cdot x_i[t]}}{\sum_{j=1}^R e^{-\beta \cdot x_j[t]}} \cdot u[t] + w[t], \\
    &\hspace{5mm} \forall \hspace{0.5mm} t \in [T], i \in [R], \\
    &A[t] \sim \mathcal{P}(x[t]),
\end{align}
where $x_i[t]$ denotes the traffic flow on each link $i \in [R]$, $u[t] \in \R$ and $w[t] \in \R$, and $\Thorizon$, are the input, zero-mean noise terms, and time horizon, as before. Here, we set $T = 250$, $\mu(0) = 0.3$, $\mu(1) = 0.2$, $u(t) = 100 R$ for each $t \in [T]$, and we again draw $w[t]$ i.i.d. from the continuous uniform distribution on $(-10, 10)$. As with the single-link case, we created two datasets for the null and alternative hypotheses. For the null hypothesis, we fix $\mathcal{P}(x[t])$ to be Bernoulli($0.02$); for the alternative hypothesis, we set $\mathcal{P}(x[t])$ to be Bernoulli($0.02$) when $x[t] < 105$, and Bernoulli($0.30$) when $x[t] \geq 105$. Again, this setting encodes the situation where higher traffic loads cause higher accident probabilities.

\begin{figure}
    \centering
    \includegraphics[scale=0.08]{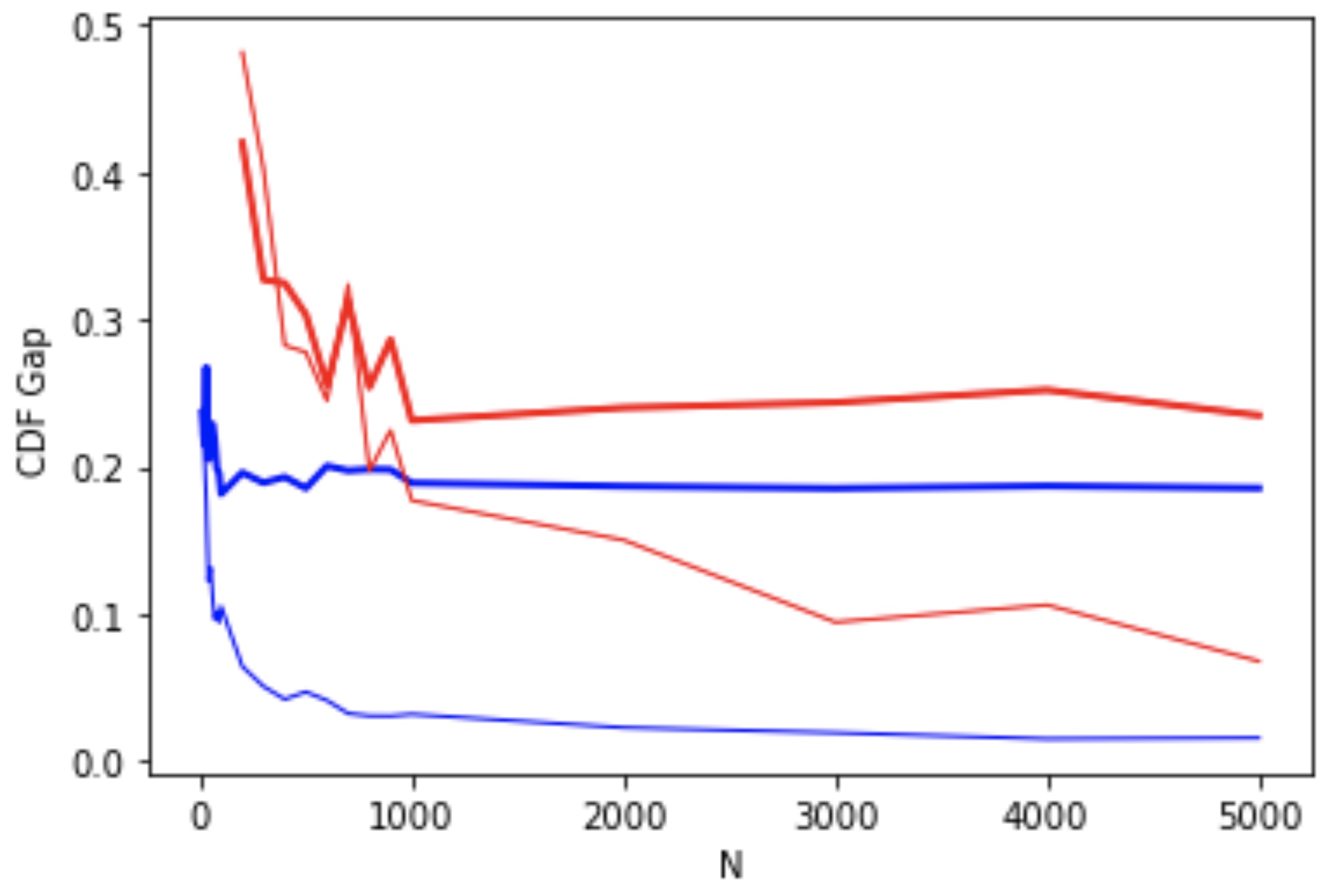}
    \caption{ \small
    CDF Gap between vs. $N$, for the 2-link traffic network example.
    Here, red and blue correspond to the baseline and our method, respectively, while thick and thin lines correspond to the null and alternative hypotheses, respectively. 
    Our approach correctly identifies the null hypothesis dataset with a relatively small number of samples, while the naive aggregation method fails to do so (thin blue curve).}
    \label{fig: CDF_Error_vs_N___2_link}
\end{figure}

Similar to the single-link case, we compute the maximum CDF gap $\sup_{x \in \R^n}|\hat b_1^N(x) - \hat b_2^N(x)|$ as functions of $N$ (thin lines), and the empirical CDFs of $X_{t-1}|T=t$ and $X_{t-1}$ (thick lines) for both the null and alternative hypotheses. We again observe that our method distinguishes between the two hypotheses at a smaller sample number $N$ compared to the baseline method.

\begin{figure}
    \centering
    \includegraphics[scale=0.08]{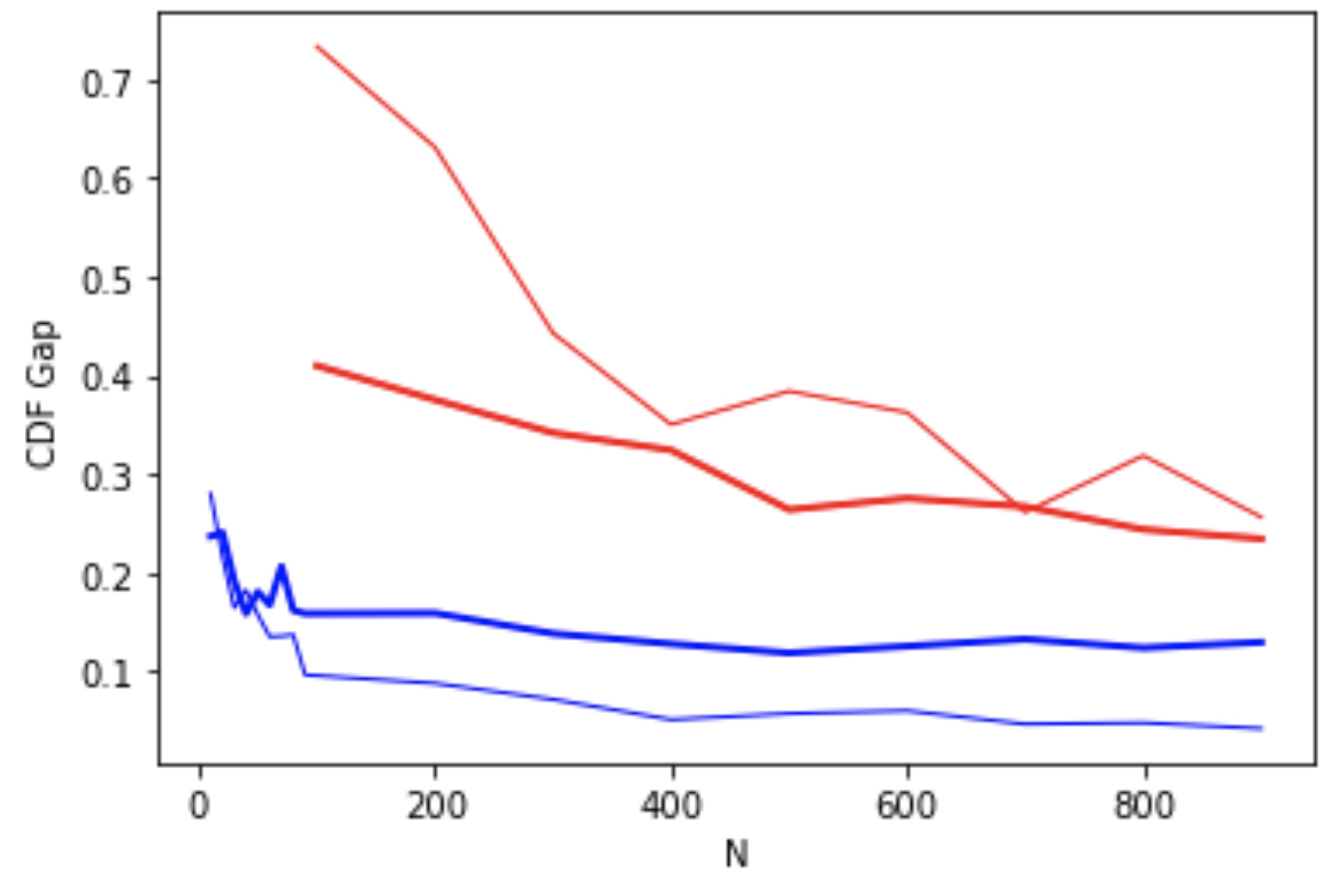}
    \caption{ \small
    CDF Gap between vs. $N$, for the 3-link traffic network example. The color and thickness schemes are identical to those of the single-link and 2-link plots in Figures \ref{fig: CDF_Error_vs_N___single_link} and \ref{fig: CDF_Error_vs_N___2_link}.}
    \label{fig: CDF_Error_vs_N___3_link}
\end{figure}

Analogous results hold for a $3$-link system with dynamics as given by \eqref{Eqn: Multi-link Traffic Network} and are presented in Figure \ref{fig: CDF_Error_vs_N___3_link}.

\end{document}